# Video Smoke Detection Based on Deep Saliency Network

Gao Xu[a], Yongming Zhang[a], Qixing Zhang[a,*], Gaohua Lin[a], Zhong Wang[b], Yang Jia[c], Jinjun Wang[a]

[a]State Key Laboratory of Fire Science, University of Science and Technology of China, Hefei 230026, China

[b]School of Computer Science and Technology, University of Science and Technology of China, Hefei 230026, China

[c]XI'AN University of Posts & Telecommunications, Xian 710121, China

Corresponding author: Qixing Zhang, Email: qixing@ustc.edu.cn, Tel: 86-551-63600770

**Abstract**

Video smoke detection is a promising fire detection method especially in open or large spaces and outdoor environments. Traditional video smoke detection methods usually consist of candidate region extraction and classification, but lack powerful characterization for smoke. In this paper, we propose a novel video smoke detection method based on deep saliency network. Visual saliency detection aims to highlight the most important object regions in an image. The pixel-level and object-level salient convolutional neural networks are combined to extract the informative smoke saliency map. An end-to-end framework for salient smoke detection and existence prediction of smoke is proposed for application in video smoke detection. The deep feature map is combined with the saliency map to predict the existence of smoke in an image. Initial and augmented dataset are built to measure the performance of frameworks with different design strategies. Qualitative and quantitative analysis at frame-level and pixel-level demonstrate the excellent performance of the ultimate framework.

**Keywords**
Video smoke detection; deep saliency network; salient map; existence prediction

# 1. Introduction

Video smoke detection is a promising fire detection method especially in open or large spaces and outdoor environments. The typical traditional video smoke detection methods consist of dynamic texture [1], wavelet based method [2], higher order linear dynamical system [3], histogram-based smoke segmentation [4], and local extremal region segmentation [5], etc. As for many other computer vision tasks, recent advances in approaches based on deep networks have resulted in significant performance improvement on the public benchmark datasets. Related work has been focused on the video smoke detection method using deep networks. Hu [6] proposes a spatial-temporal architecture for a multitask learning to recognize smoke and estimate optical flow simultaneously. Luo [7] introduces a smoke detection algorithm based on the combination of motion characteristic and convolutional neural network (CNN). Xu designs a deep domain adaptation network in [8] for smoke recognition using synthetic smoke images. This paper introduces a novel approach for smoke recognition based on saliency detection. To the best of our knowledge, we are the first to investigate smoke detection based on deep saliency network.

Visual saliency detection, which aims to highlight the most important object regions in an image, has been a fundamental problem drawing extensive attentions in recent years. It has been widely utilized for many computer vision tasks, such as semantic segmentation, object tracking, and image retrieval, etc.

As reviewed in [9], the numerous salient object detection methods have been proposed in recent years. Overall, saliency detection methods can be divided into traditional methods and CNN based methods. In [10], Jiang uses the supervised learning approach based on multi-level image segmentation to map the regional feature vector to a saliency score, and finally fuses the saliency scores across multiple levels to yield the saliency map. Zhu [11] proposes to generate initial saliency map based on color and depth saliency map, and center-dark channel map based on center saliency prior and dark channel prior, respectively. Then these saliency maps are fused to generate the final saliency map.

Since the latest generation of CNNs have substantially outperformed handcrafted approaches in computer vision, CNN based saliency detection methods have attracted wide attention for their superior performances. CNN based saliency detection methods mainly include region-level saliency detection methods and pixel-level saliency detection methods. For the region-level methods, Li [12] proposes to incorporate multiscale CNN features extracted from nested windows with a deep neural network with multiple fully connect layers. Zhao [13] designs a multi-context deep models on the superpixels of an image, including global-context modeling for saliency detection with a superpixel-centered window and local-context modeling with a closer-focused superpixel-centered window. For the pixel-level methods, Cheng [14] proposes a deeply supervised salient object detection method (DSS) by introducing short connections to the skip-layer structures within the holisitcally-nested edge detector (HED) architecture. Li [15] proposes a multi-task model (DS) based approach to model the intrinsic semantic properties of salient objects and presents a fine-grained super-pixel driven saliency refinement model for the output of the model of the proposed fully convolutional neural network (FCNN). Deep hierarchical saliency network (DHSNet) [16] adopts a novel hierarchical recurrent convolutional neural network (HRCNN) to refine saliency maps in details by incorporating local contexts.

More and more deep networks focus on the combination of multiple branches for information fusion, such as extra features for CNN-based detector in [17], simultaneous detection and segmentation in [18]. The typical application is the combination of handcrafted features and deep CNN features, e.g. Li [19] integrates handcrafted low-level features with deep contrast features for a more robust feature. Qu [20] designs a CNN to automatically learn the interaction between the different low-level saliency cues and takes advantage of the knowledge obtained in these traditional saliency detections. In this paper, an end-to-end framework for salient smoke detection and existence prediction of smoke is proposed for application in video smoke detection. The pixel-level and object-level salient CNNs are combined to extract the informative smoke saliency map. Experiment results show that our method outperforms the state-of-art salient object detection methods for video smoke detection.

## 2. Related work

There are many works using multiple branches for saliency detection. Chen [21] proposes a saliency model built upon two stacked CNNs. The first CNN generates a coarse-level saliency map in the global context. The second CNN integrates superpixel-based local context information in the first CNN to refine the coarse-level saliency map. Li [22] designs a deep contrast network (DCL) consisting of pixel-level fully convolutional stream and a segment-wise spatial pooling stream. Tang [23] proposes a saliency detection method by combining region-level saliency estimation and pixel-level saliency prediction with CNNs (CRPSD). The first stream produces a saliency map in pixel-level based on deeplab [24], and the second stream extracts segment-wise features. Finally, a proposed fully connected conditional random field (CRF) model is optionally incorporated to refine the fused result from these two streams. Qu [20] develops a graph Laplacian regularized nonlinear regression scheme for saliency refinement to generate a fine-grained boundary-preserving saliency map. There are some technologies of spatial coherence refinement for the post-processing approaches, such as CRF [19, 22, 24] , clustering [12, 21] using superpixel, graph Laplacian regularized nonlinear regression [15, 20]. Zhao [13] outputs the occupation ratio of salient softmax value, while the general saliency network output sigmoid of probability of saliency.

However, most of these methods infer saliency by learning contrast through the existing fully convolutional network in VGG [25] network, and the global context information is not used preferentially. A standard fully convolutional model is not particularly good at capturing subtle visual contrast in an image. In addition, in the two-stream baselines CRPSD [23] and DCL [22] for the combination of pixel-level and region-level saliencies, the region-level saliency maps make a certain effect to the final saliency map as region-level saliency detection enhances the edge of the salient object. As the edge of smoke is fuzzy, the region-level saliency detection of smoke is not well compared to the general object. Therefore, our method use the recurrent connections in our pixel-level CNN to propagate the global context information to local context hierarchically and progressively. The object-level saliency map generated from objectness is used to generate the final smoke saliency map fused with pixel-level saliency map, as the objectness can offer the useful location information of smoke pixels.

In summary, this paper has the following contributions:

- We investigate the performance differences between state-of-art saliency detection methods for smoke detection.
- We propose an end-to-end framework for salient smoke detection and prediction for smoke existence.

- The integration of multiple-level saliency cues is proposed for smoke detection, and a detailed analysis of the time consuming and performance is provided.
- Qualitative and quantitative evaluations at frame-level and pixel-level demonstrate the excellent performance of the proposed method.

The rest of this paper is organized as follows. In Section 3, we give an overview of the multiple level saliency detection methods for smoke detection. Then the proposed salient smoke detection method is introduced in Section 4. Section 5 gives the experimental analysis for measurement. The conclusion and future work are presented in Section6.

## 3. Object-level and region-level saliency detection

We introduce the different level saliency detection methods for smoke detection, including the object-level and region-level saliency detection. The produced object-level and region-level saliency map will be fused with pixel-level saliency map respectively.

**3.1 Object-level saliency detection**

Object detection can make contribution to segmentation in some works [17, 26]. As introduced in [27], the object proposal algorithms can measure how likely a region contains an object without the need of category information, which can be used to obtain high-level saliency priors. There is a similar case [28] that objectness heatmaps are obtained from the proposals generated by region proposal network (RPN) [29]. Typically, the graph-based proposal method - selective search segmentation [30] can't recognize the smoke region well, compared to the recognition of the general rigid object as shown in Fig. 1. We use the RPN to generate the object-level smoke saliency map. Our RPN model is trained on the dataset with bounding box annotation for smoke detection. RPN outputs the candidate boxes with confident scores and the objectness score of each pixel is normalized to [0, 255]. We follow the way for the objectness calculations and employment in [27]. For each bounding box $B_i$ with the confidence score $b^i$ generated by RPN model in an image, its confidence score $b^i$ is added to all the pixels in the bounding box. The objectness score $s_p$ of each pixel in the image is computed as the square root of the summed squares of the confidence scores from all the

bounding boxes that covers it, weighted by a Gaussian function for smoothness :

$$s_p = [\sum_{i=1}^{N} b_i^2 I(p \in B_i) \exp\{-\lambda d(p, B_i)\}]^{1/2} \qquad (1)$$

Where $d(p, B_i)$ is the normalized distance between the pixel $p$ and the center of the bounding box $B_i$ (total number is $N$). $I(p \in B_i)$ indicates whether $p$ is inside the $B_i$. As shown in Fig. 2, a relatively accurate location information of smoke can be provided in the heatmap obtained from the objectness score of each pixel. For reference, the visualization of feature maps on Conv4_3 and Conv5_3 in RPN is also given.

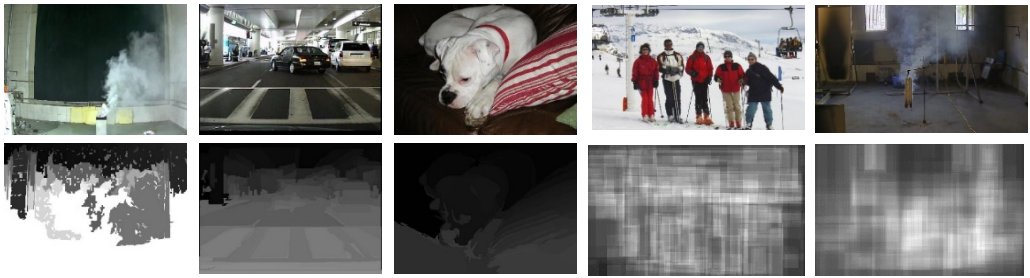

Fig. 1. The top row represents the original images, the bottom row represents the initial segmentation results and the heatmap of selective search method.

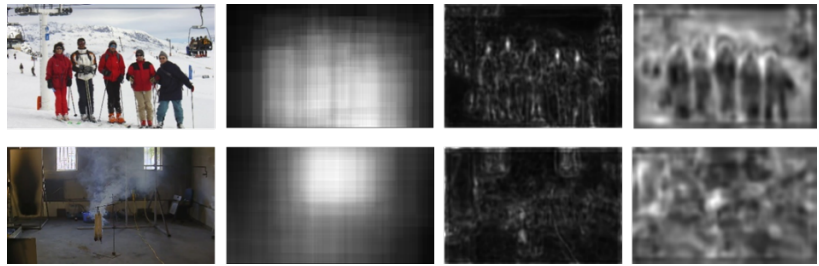

Fig. 2. From left to right are: original image, heatmap, feature map on Conv4_3, feature map on Conv5_3.

**3.2 Region-level saliency detection**

Region-level segmentation extracts features of regions in an image as context to predict saliency score of each region, typically as superpixels based approach. The superpixel segmentation is performed using the simple linear iterative clustering (SLIC) algorithm [31], which uses geodesic image distance during K-means clustering in the CIELab color space. As shown in Fig. 3, all pixels in each superpixel are in the same gray value. For visualization, superpixels with half of the pixels overlap with the smoke region are in white, and the small black blocks represent the center of each superpixel. As analyzed in [23], if there are too few superpixels, the smoke region will be under-segmented; if there are too many superpixels, the smoke region or background will be over-segmented. Fig. 3 shows the segmentation results with different number of superpixels. Here, we set the number of superpixels as 100 considering both of effectiveness and efficiency. Since the superpixels based approach [13] gets the state-of-art performance, we set it as region-level saliency CNN to generate the region-level smoke saliency maps, as shown in the bottom row of Fig. 3.

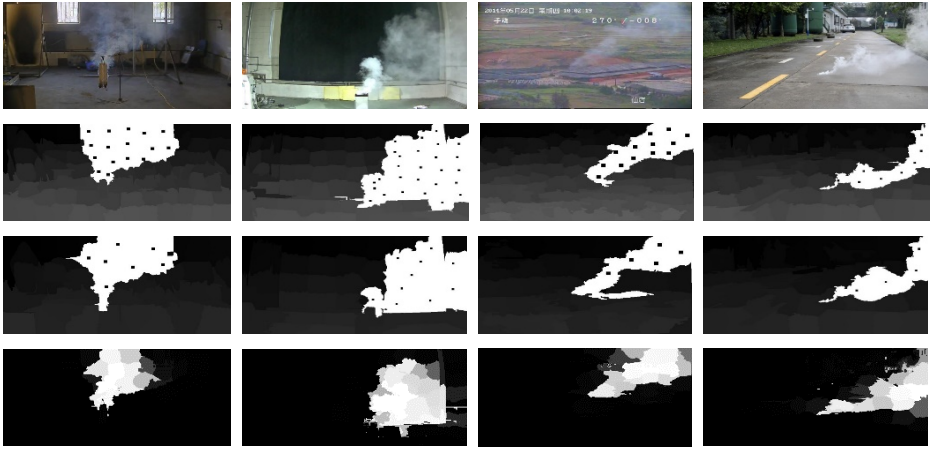

Fig. 3. The superpixels with half of the pixels overlap with the smoke region are in white for visualization and the small black blocks represent the center of each superpixel. The top row represents the original images, the second row represents the segmentation with 100 superpixels, the third row represents the segmentation with 80 superpixels and the bottom row represents the region-level saliency maps.

The object-level and region-level smoke saliency map will be used to the combination with the pixel-level smoke saliency map in our framework. Considering the effectiveness of the approach, time-consuming statistics are given in Table 1.

Table 1. Time-consuming statistics in the test procedure. The stage 1 and stage2 are the global-context stream and local-text stream in the superpixels based method [13] respectively. The fusion method runs based on the computed object-level or region-level saliency maps. Obviously, the SLIC algorithm is time-consuming.

|  | SLIC | RPN | Stage1 | Stage2 | Fusion |
| --- | --- | --- | --- | --- | --- |
| Time (s) | 0.4804 | 0.0578 | 0.6994 | 0.5982 | 0.0166 |

## 4. Architecture

As shown in Fig. 4, the architecture of the proposed method consists of three complementary components,

an encoder-decoder architecture based on fully convolutional network, the object-level and region-level saliency CNN and the existence prediction branch.

**4.1 Pixel-level saliency CNN**

As shown in Fig. 4, the encoder architecture of the framework is built based on the 13 convolutional layers (from layer Conv1_1 to Conv5_3) in VGG16 [25]. Each convolution layer owns $3 \times 3$ kernel. The first 3 pooling layers have stride of 2 and $2 \times 2$ kernel, while the 4[th] pooling layer has a stride of 1 and $3 \times 3$ kernel. The top feature map of the base network is of $8 \times$ subsampled resolution. Inspired by DHSNet [16], the recurrent convolutional layer [32] (RCL) is used to incorporate local contexts efficiently in feature maps in encoder (Conv1_2, Conv2_2, Conv3_3, Conv4_3) for refining the detail of saliency map in the decoder (SmRCL1, SmRCL2, SmRCL3, SmRCL4) . The detailed architecture of recurrent convolution layer is shown in Fig. 5. A convolution layer with $1 \times 1$ kernel follows the top feature layer Conv5_3. The 3 deconvolutional layers in the decoder architecture own stride of 2 and $2 \times 2$ kernel, and the final pixel-level saliency map has the same resolution with the original input image.

**4.2 Saliency map fusion**

As descripted in Section 3, the object-level saliency map is generated through proposals produced by the RPN, and the region-level saliency map is generated through the superpixels based approach [13]. The pixel-level, object-level and region-level saliency maps are obtained by using different information of images. The saliency maps are concatenated to the maps X, and their complementary information are combined using the nonlinear manner in a convolutional layer with $1 \times 1$ kernel. Then a sigmoid activity function is used to produce the final prediction,

$$U = \frac{1}{1 + e^{-f(X)}} \qquad (2)$$

Where $f(X) = W \times X + b$ is the convolution operation.

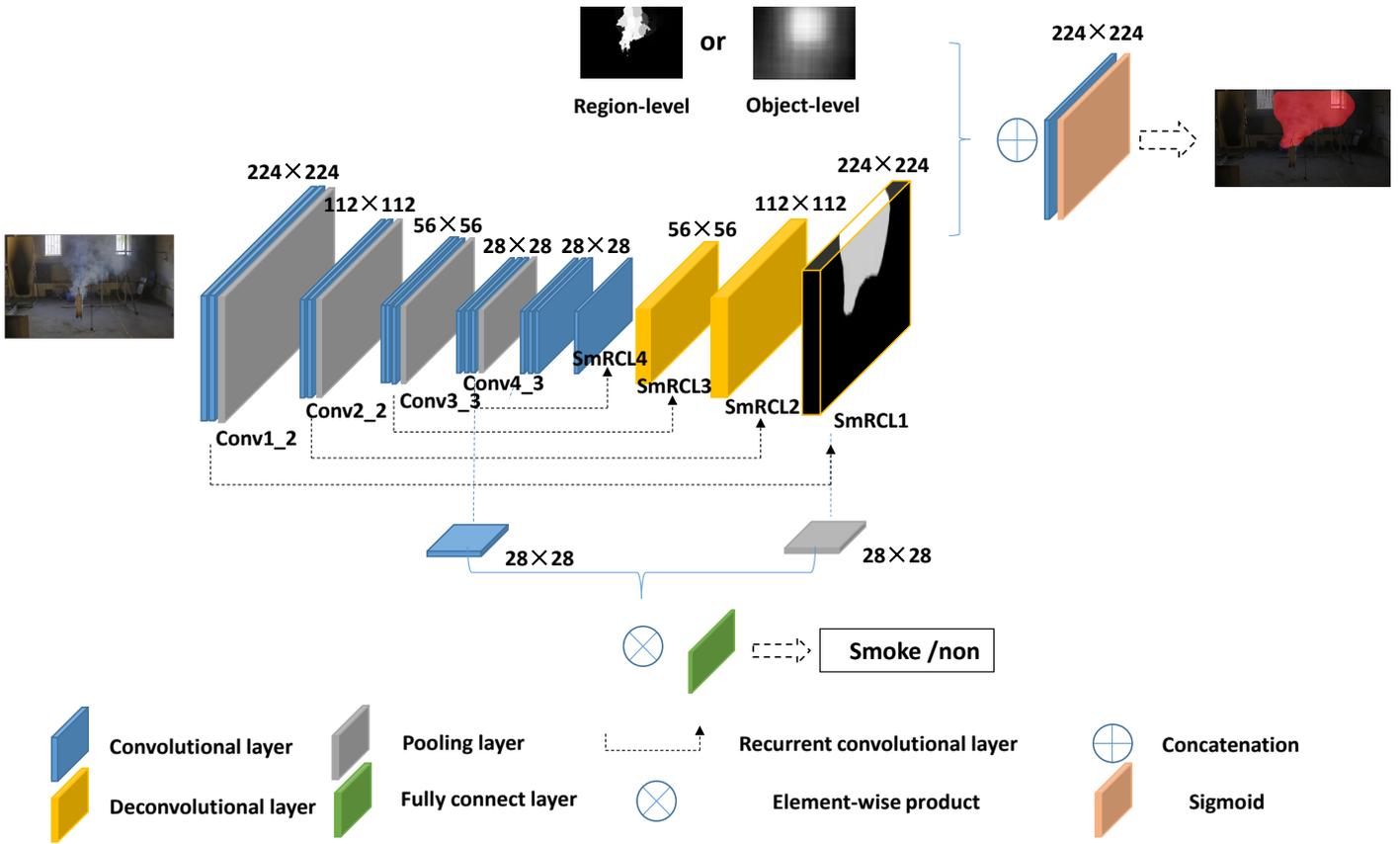

Fig. 4. The overall framework consists of the master branch (pixel-level CNN) and partner branch for existence prediction. The final smoke saliency map is the fusion of pixel-level and object-level saliency maps through experiment comparison. The feature map and saliency map are combined for existence prediction of smoke.

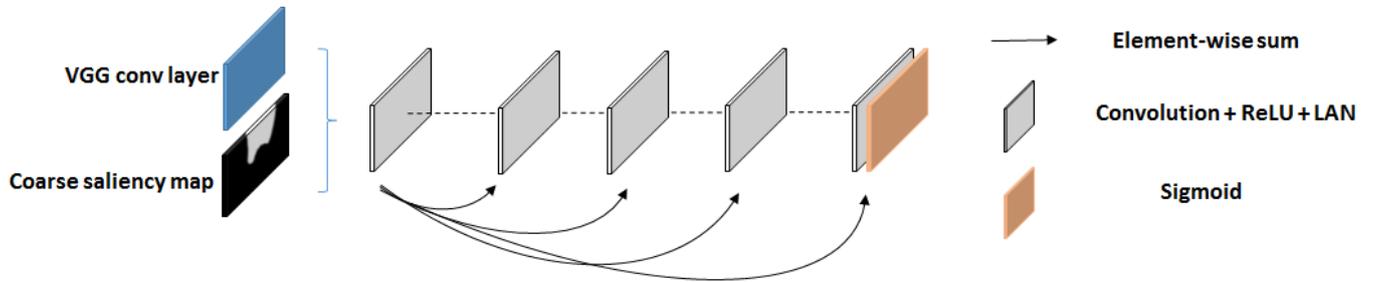

Fig. 5. The overview of the framework of the RCL [32] (recurrent convolutional layer). The first 4 convolutional layer own $3 \times 3$ kernel, while the 5th convolutional layer owns $1 \times 1$ kernel.

In the experiments, we found that the combination of object-level and pixel-level saliency map performs better than the combination of region-level and pixel-level saliency map. In addition, the generation of superpixels for the region-level saliency detection is time consuming (as shown in Table 1), so we choose to fuse the object-level and pixel-level saliency map in the ultimate framework.

### 4.3 Existence prediction

The other task of our approach is to predict the existence of smoke in an image. As shown in Fig. 6, sometimes the output of the saliency prediction is some discrete pixels or areas.

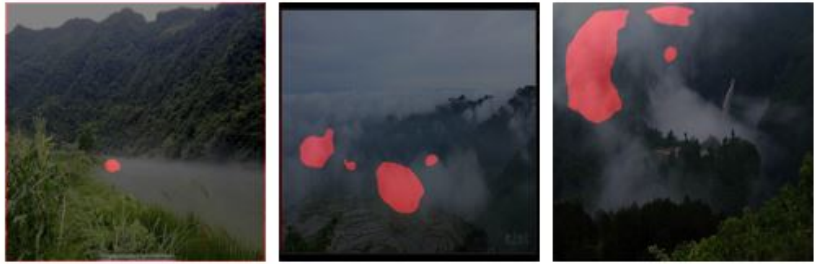

Fig. 6. The saliency map in the hard samples.

In this case, it is difficult to judge whether smoke exists in the image. So we propose to integrate the existence prediction branch to the framework. The existing saliency detection methods assume that at least one salient object exists in the input image. As smoke appears with low probability, there are many background-only images in our dataset. As shown in Fig. 4, the partner CNN branch in the bottom of the framework is used to predict smoke existence through the combination of the feature map in encoder architecture and the saliency map in decoder architecture. As the saliency map offers the probability map of smoke, and the feature map can give the informative representation of the entire image, the combination of them is able to extract the feature of high salient candidate to predict the existence of smoke. We design variants with the following 6 strategies for the structure of the existence prediction branch as follows.

Strategy 1: A fully connect layer is set following the highest level saliency layer SmRCL1.

Strategy 2: 2 convolutional layers (kernel_size: $3 \times 3$, num_output: 64) and 1 fully connect layer are set following the SmRCL1.

Strategy 3: A average pooling layer (kernel_size: $8 \times 8$, stride: 8) is set following the SmRCL1, and a convolutional layer (kernel_size: $1 \times 1$, num_output: 1) is set following the highest level feature layer Conv4_3. These two outputs are combined through an elsewise-wise product operation, and a fully connect layer is then set.

Strategy 4: A average pooling layer (kernel_size: $8 \times 8$, stride: 8) is set following the SmRCL1, and a convolutional layer (kernel_size: $1 \times 1$, num_output: 1) with a sigmoid activation function is set following the highest level feature layer Conv4_3. These two outputs are combined through an elsewise-wise product operation, and a fully connect layer is then set.

Strategy 5: A spatial pooling layer (kernel_size: $8 \times 8$, stride: 8) and 1 fully connect layer are set following the conv4_3.

Strategy 6: A convolutional layers (kernel_size: $1 \times 1$, num_output: 1) is set following the layers Conv1_2. The output is combined with the saliency layers SmRCL1 through element-wise product operation, then 1 fully connect layer is set.

The overall loss function of the framework consists of the softmax loss on frame-level prediction and cross-entropy loss on pixel-level prediction. Giving a training example $(E, Y_f, Y_p)$ with frame-level label $Y_f$ and pixel-level label $Y_p$,

$$L(Y_f, Y_p) = -\sum_{i}^{h \times w} (\alpha \times \log P(Y_p = 1 \mid p_i) + (1 - \alpha) \times \log P(Y_p = 0 \mid (1 - p_i))) - \sum_{j}^{2} Y_f \log(z_j) \quad (3)$$

Where, $\alpha$ means the ratio of salient pixels in ground truth $Y_p$, $p_i$ represents the saliency value of the pixel, $h$ and $w$ are the height and width of the image respectively, and $z_j$ is the softmax value of the existence prediction.

## 5. Experiments

### 5.1 Evaluation metrics

We evaluate the performance using precision-recall (PR) curve and F-measure. Precision corresponds to the percentage of salient pixels correctly assigned, while recall corresponds to the fraction of detected salient pixels in relation to the ground truth number of salient pixels. The PR curve reflects the object retrieval performance in precision and recall by binarizing the saliency map using different thresholds (ranging from 0 to 255). F-measure characterizes the balance degree of object retrieval between precision and recall. By using an image adaptive threshold, the F-measure is computed as:

$$F_\beta = \frac{(1 + \beta^2) \times precision \times recall}{\beta^2 \times precision + recall} \quad (4)$$

Where, $\beta^2 = 0.3$ as suggested in the previous work [33], and the adaptive threshold is determined to be twice the mean saliency value of each saliency map.

**5.2 Datasets**

We built an initial dataset for the comparison of our approach with the different state-of-art methods, which is available at http://smoke.ustc.edu.cn. The training set consists of 1401 smoke images and 1499 non-smoke images, while the test datasets consists of 1399 smoke images and 1401 non-smoke images. In addition, as far as we know, the data augmentation is rarely mentioned in related work of saliency detection. Compared to the public saliency dataset rich in scale and diversity, smoke images are extracted from the limited video. As smoke detection belongs to event detection, we regard the smoke detection task as salient smoke detection and smoke existence prediction. To perform a frame-level (existence prediction) and pixel-level (saliency prediction) evaluation, we train and test the framework for saliency detection and existence prediction on the augmented dataset rich in diversity. The enhanced dataset is built using these two measurement:

1. We collect the hard negative samples such as cloud and fog. Furthermore, the original smoke images are superimposed with the non-smoke images, as shown in Fig. 7.

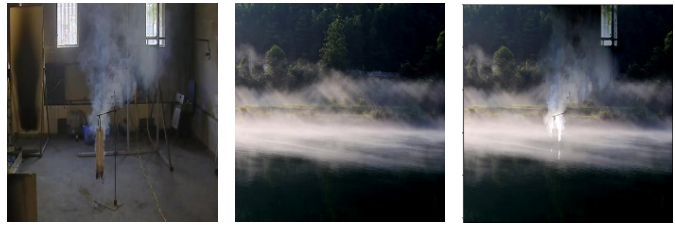

Fig. 7. The visual effects of composition using gradient domain cloning. However, it is time consuming. In the future work, we will apply some efficient method to overlay images.

2. Inspired by Hide and seek [34], the pixel-level ground-truth of smoke region is hided randomly at horizontal and vertical direction, as shown in Fig. 8. Considering the deformation and media properties of smoke, the occluded smoke images can be used to enhance the diversity of dataset.

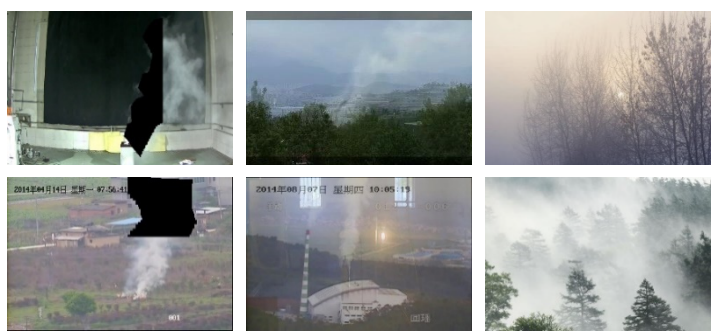

Fig. 8. The challenge images in the augmented dataset. From left to right are: the occluded smoke images, the composite smoke images, and the hard negative images.

After the augmentation, the training set consists of 5596 smoke images and 4852 non-smoke images, and the test set consists of 1399 smoke images and 1212 non-smoke images, which is also available at https://smoke.ustc.edu.cn . The model of the framework with different design choices for existence prediction branch is trained and evaluated on this dataset.

**5.3 Results**

We compared our method with other 6 state-of-art models, including FCN [35], DHSNet [16], DSS [14], CRPSD [23], DCL [22], and DS [15]. Like the experiment in DCL [22], the softmax layer in FCN (FCN-8s network) is replaced with a sigmoid cross-entropy layer for saliency inference in our experiment. In addition, the fusion model with pixel-level and object-level saliencies, and the fusion model with pixel-level and region-level saliencies in our method are evaluated respectively.

As shown in Fig. 9, the fusion model with pixel-level and object-level saliencies outperforms the other saliency model in terms of F-measure and PR curve. Meanwhile, Fig. 10 shows the visual comparison of the segmentation results generated from the 8 methods respectively. The segmentation result is generated from the saliency map binarized using the adaptive threshold according to [33].

It can be seen that the fusion model with pixel-level and object-level saliencies can generate more accurate segmentation result, and the segmentation results of FCN and MD are dispersed and broken, as the inherent spatial invariance of FCN does not take into account the useful global context information. The prediction of FCN and MD give a clear false positive on the challenging non-smoke images. MD carries out the task of saliency detection in conjunction with the task of object class segmentation, and the two tasks share the convolutional layers of FCN. Compared to FCN, the performance of smoke saliency prediction of MD is hardly improved. The performances of the other 4 saliency model DHSNet, DSS, CRPSD and DCL are clear better than that of FCN and MD. This is due to the fact that there are specially architectures for saliency detection in these model. The recurrent connect layers in DHSNet enhance the capability of the model to integrate context information for refining smoke saliency prediction, which also make contribution to our methods. The atrous convolution used in DCL helps to produce accurate smoke pixel location. CRPSD integrates the region-level saliency into the final saliency map, as the region-level information is used to model visual contrast between regions and visual saliency along region boundaries. The performance of DSS exceeds slightly the fusion model with pixel-level and region-level saliencies in our method. As DSS designs short connections between shallower and deeper side-output layers, the activation of each side-output layer gains the capability of both highlighting the entire salient smoke region and accurately locating its boundary. The fusion model with pixel-level and object-level saliencies in our method achieves the best performance. The object-level information is helpful to obtain discovery of salient smoke region more precisely. The objectness scores generated from the RPN can give a high-level saliency prior, and the combination of information of pixel-level and object-level saliencies can further improve the performance.

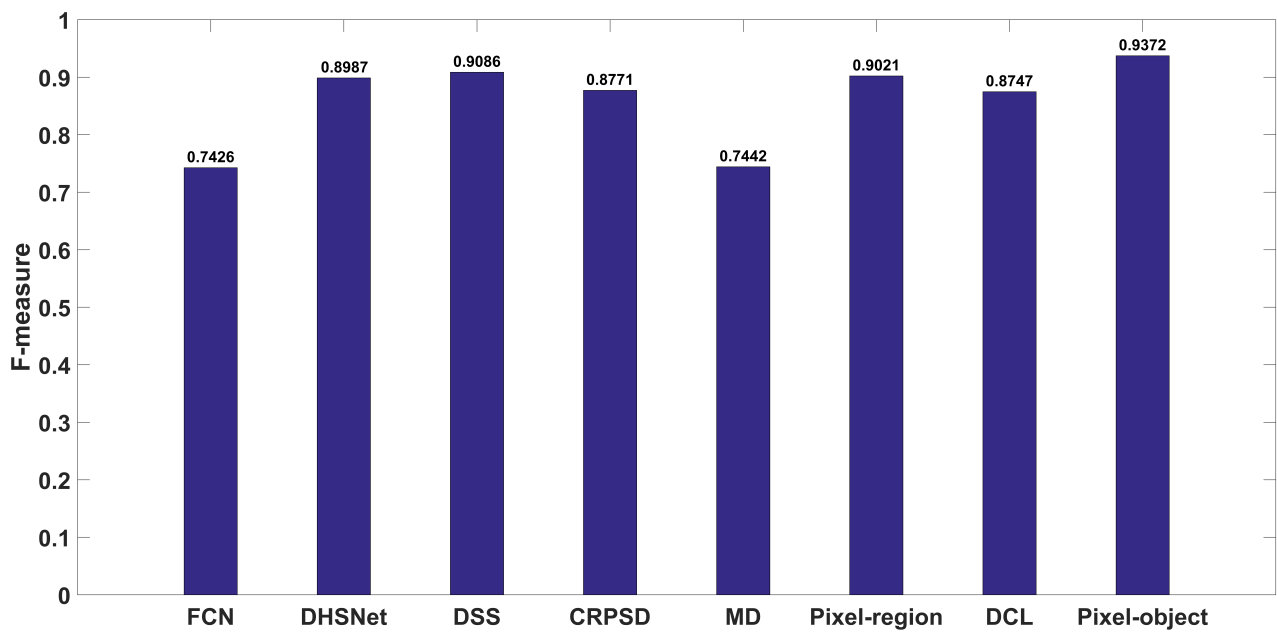

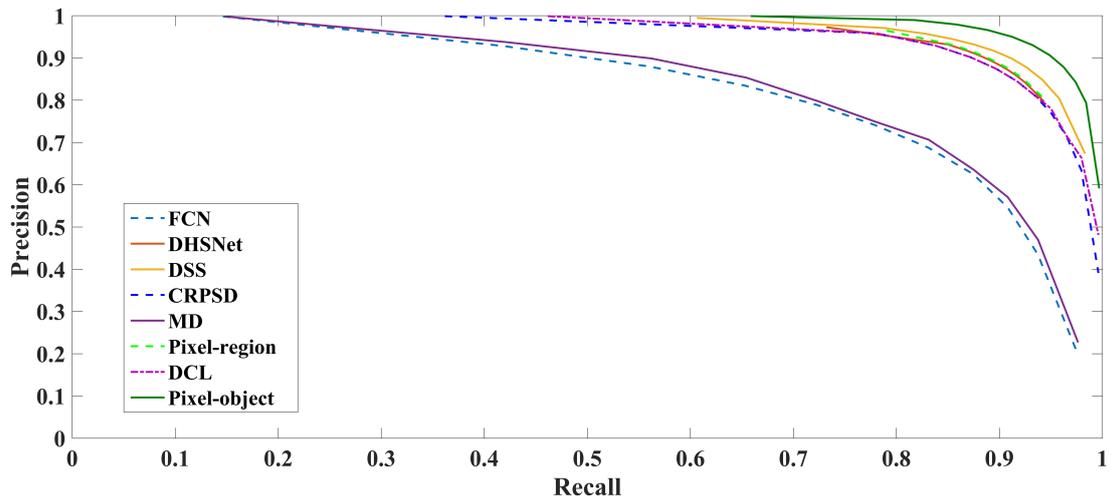

Fig. 9. Comparison of the the values of F-measure and precision-recall curves of the 8 saliency detection methods for smoke detection, including FCN, DHSNet, DSS, CRPSD, MD, fusion model with pixel-level and region-level saliencies, DCL and fusion model with pixel-level and object-level saliencies.

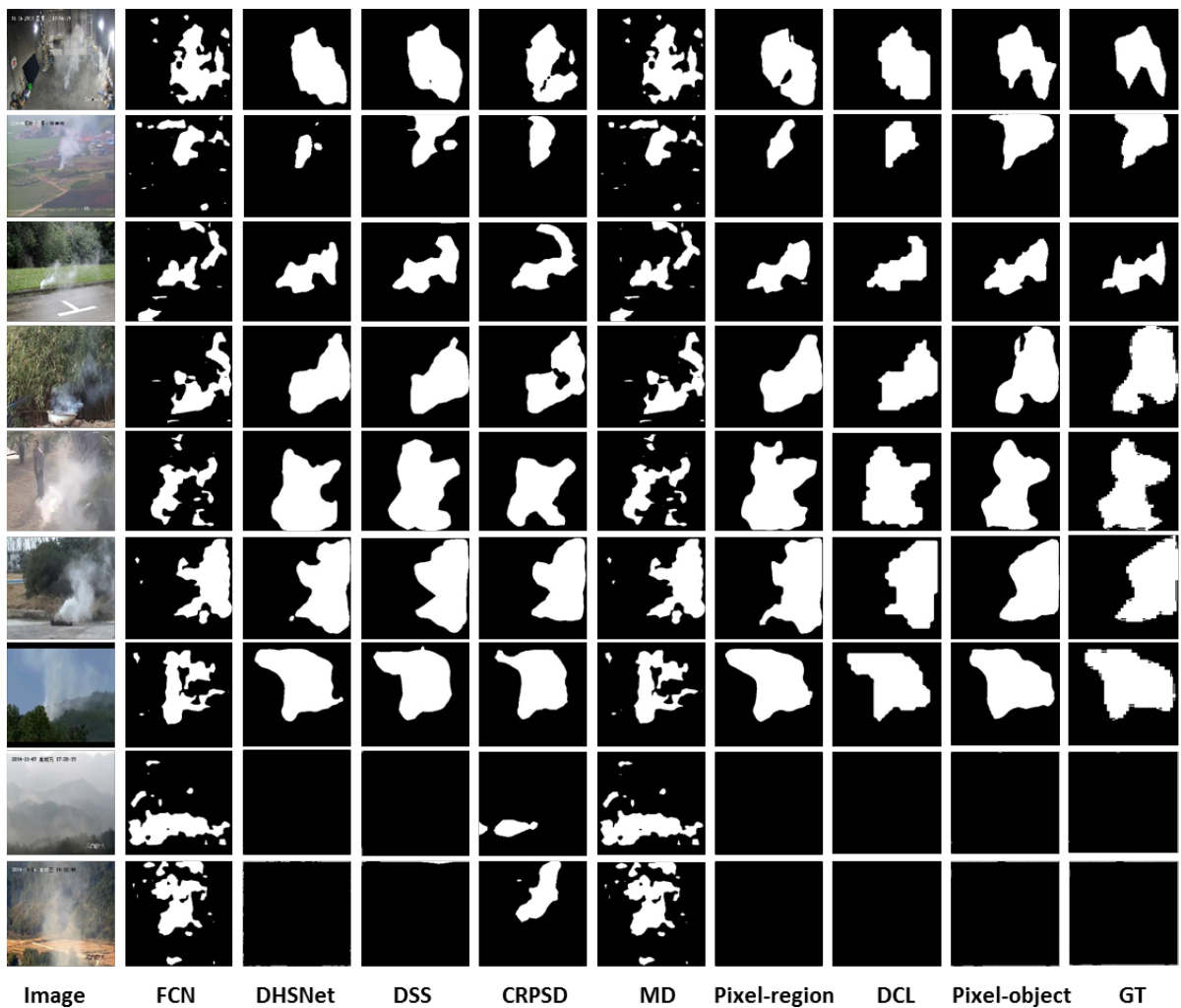

Fig. 10. Visual comparison of segmentation results from the 8 methods. The ground truth (GT) is shown in the last column. The segmentation result is generated from the saliency map binarized using the adaptive threshold according

to [33].

In [36], the local and global color contrast, local gPB boundary strength and object size are used to measure the dataset bias which causes the degradation of performance of salient object detection. In our work, we analyze the following image statistics in order to find the interaction between performances of saliency models and image characteristics, including RGB histrom contrast (the Chi-square distance between the smoke region RGB histogram and background RGB histogram), size of smoke region (the area ratio of smoke region to background), thickness of smoke region (the ratio of the total gray value to the area of smoke region) and dispersion of smoke (the variance of the distributions of pixel locations of smoke region).

As shown in Fig. 11, it can be seen that while the value of region-size and dispersion of smoke region get smaller , the overlap value of segmentation result of each model and ground-truth annotation drops, although the value of thickness gets larger. While the value of region-size and dispersion of smoke region get larger, the overlap value of segmentation result of each model and ground-truth annotation rises, although the value of RGB histrom contrast gets smaller. It can be seen that the performance of detector is more related to the region size and dispersion than the RGB histrom contrast and thickness.

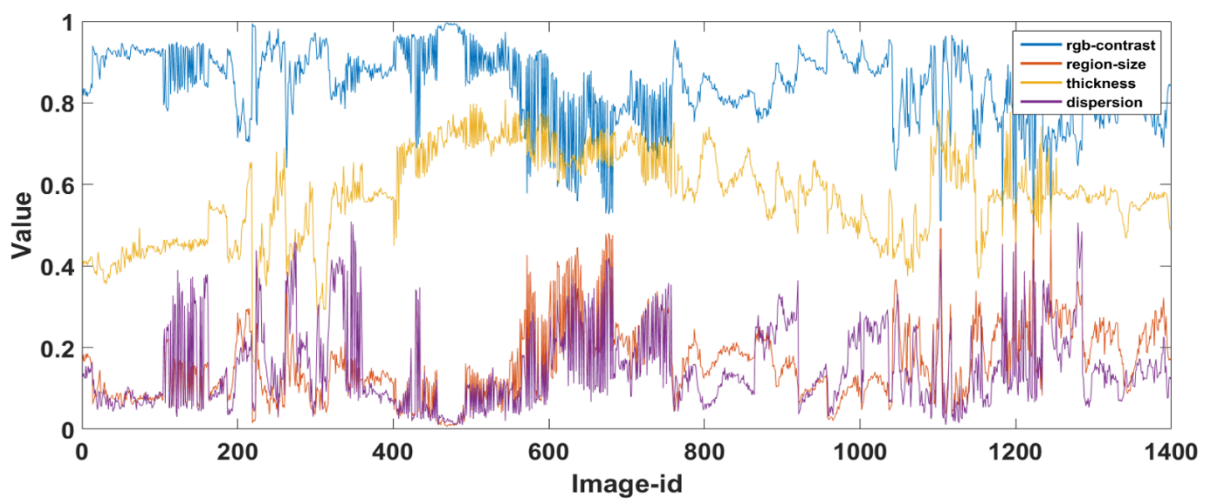

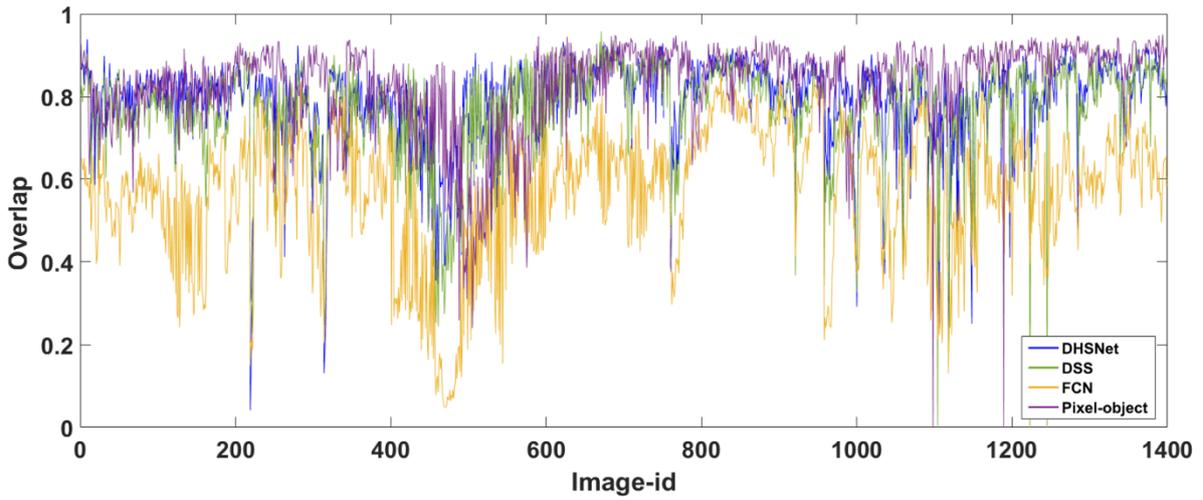

Fig. 11. The top figure represents the image statistic value of each smoke image in test set. And the bottom figure represents the overlap value of segmentation result of each model and ground-truth annotation in each smoke image of test set.

On the other hand, we train and test the model of the framework for saliency detection and existence prediction. In can be seen in Fig. 12 that the different design choices of the existence prediction branch make little effects on the saliency prediction results, but the performances on frame-level prediction of smoke existence is reversed. The model designed with strategy 3 achieves the best performance in the accuracy of existence prediction. As the deep feature map in Conv4_3 gives more semantic information about smoke region and the top saliency map in SmRCL1 integrates the fine detail with the global and local information, the combination of them can extract the features of high salient candidate to predict the existence of smoke.

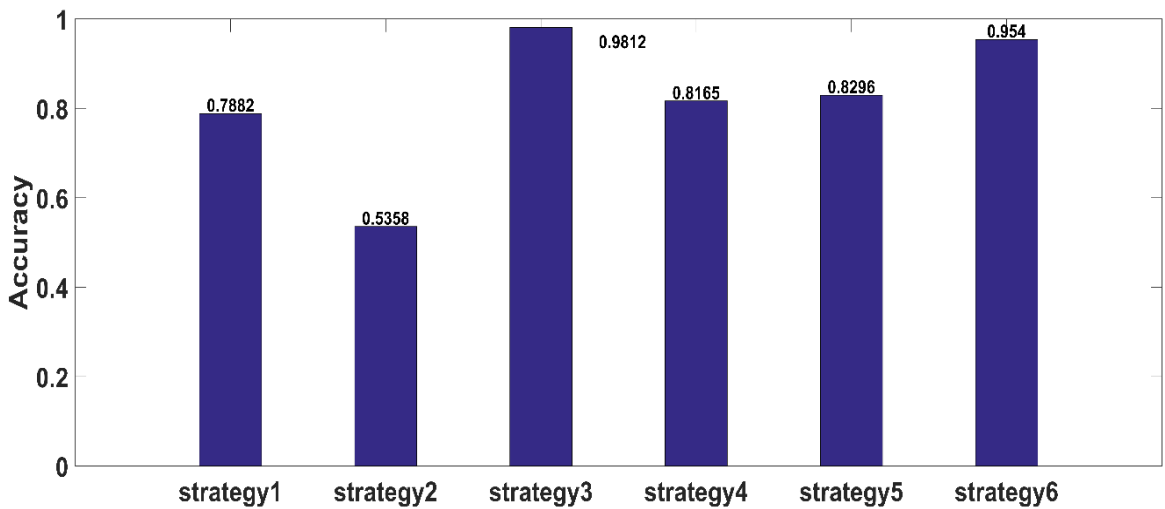

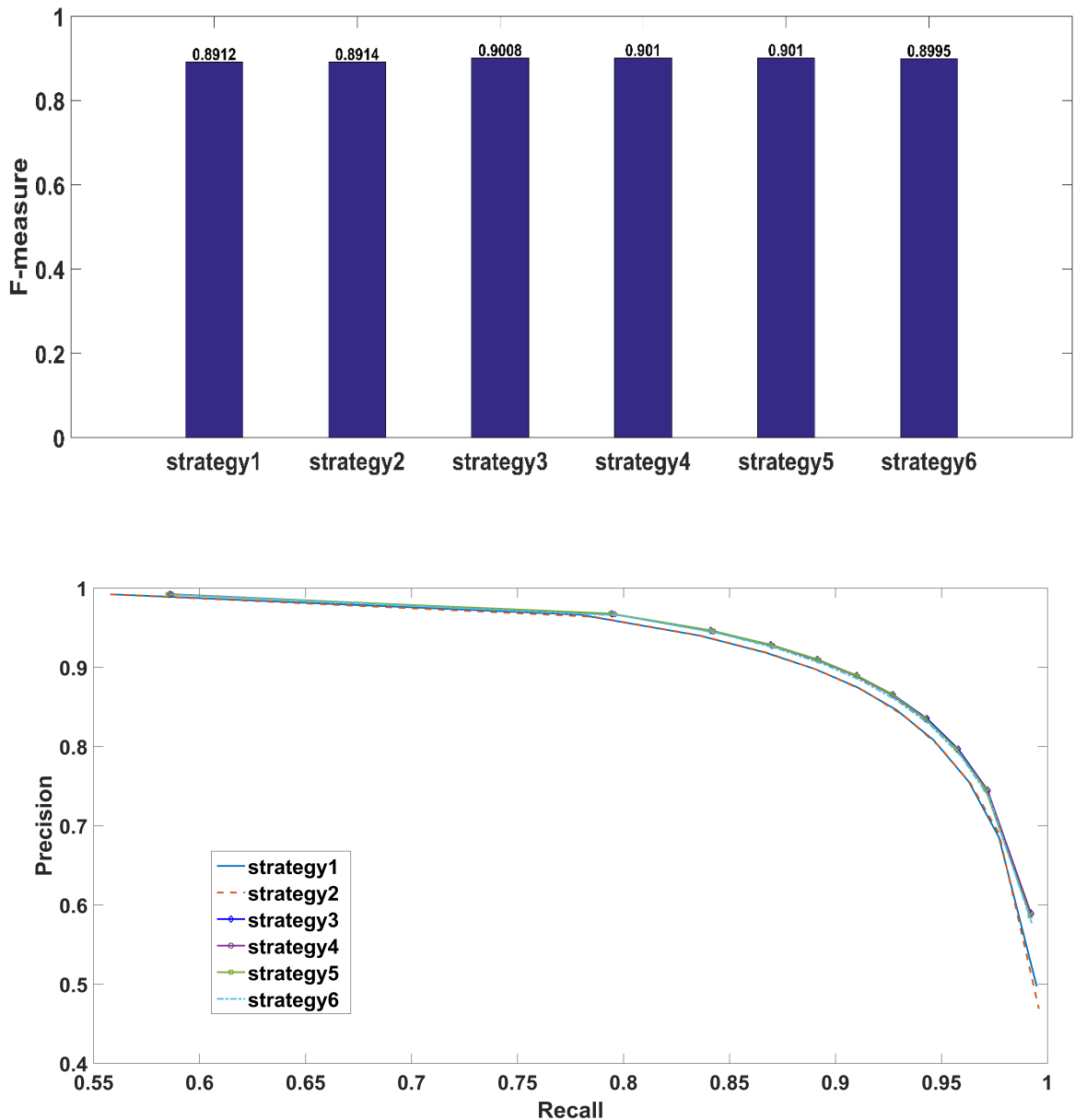

Fig. 12. From top to bottom: the existence prediction accuracy, the F-measure scores and PR curves. We present the performance of the joint network with different design choices.

## 6. Conclusion

For video smoke detection, we systematically compare several state-of-art saliency detection methods, including handcraft-feature and CNN based methods. The pixel-level, object-level and region-level salient CNN are combined to extract the informative smoke saliency map. The region-level salient CNN is abandoned as the combination of object-level and pixel-level saliency map performs better than the combination of region-level and pixel-level saliency map and the generation of superpixels for the region-level salient CNN is time consuming. An end-to-end framework for salient smoke detection and existence prediction is proposed for the application in video smoke detection. Qualitative and quantitative evaluations at frame-level (existence prediction) and pixel-level (saliency prediction) demonstrate the excellent performance of the ultimate framework. In the future, the dataset richer in diversity and complex in scene will be created and more efforts on salient smoke detection in video will be put to characterize the smoke saliency in temporal-spatial.

**Acknowledgements**

This work was supported by the Anhui Provincial Key Research and Development Plan under Grant No. 1704a0902030, the National Key Research and Development Plan under Grant No. 2017YFC0805100 and 2016YFC0800100, and the Fundamental Research Funds for the Central Universities under Grant No. WK2320000040. The authors gratefully acknowledge all of these supports. The authors specially thank Professor Zhiqiang Zhou for sharing the wildfire smoke dataset with us.